\pdfoutput=1

\documentclass[11pt]{article}

\usepackage{emnlp2021}

\usepackage{times}
\usepackage{latexsym}
\usepackage{graphicx}
\usepackage{amsmath}
\usepackage{color}
\usepackage{multirow, tabularx}
\usepackage{subcaption}

\usepackage[T1]{fontenc}

\usepackage[utf8]{inputenc}

\usepackage{microtype}

\title{Multiplex Graph Neural Network for Extractive Text Summarization}

\author{Baoyu Jing$^\dag$, Zeyu You$^\ddag$, Tao Yang$^\ddag$, Wei Fan$^\ddag$ \and Hanghang Tong$^\dag$ \\
        $^\dag$University of Illinois at Urbana-Champaign\\
        $^\ddag$Tencent America\\
        \{baoyuj2, htong\}@illinois.edu\\
        \{zeyuyou, tytaoyang, davidwfan\}@tencent.com}

\begin{document}
\maketitle
\begin{abstract}
Extractive text summarization aims at extracting the most representative sentences from a given document as its summary. 
To extract a good summary from a long text document, sentence embedding plays an important role.
Recent studies have leveraged graph neural networks to capture the inter-sentential relationship (e.g., the discourse graph) 
to learn contextual sentence embedding.
However, those approaches neither consider multiple types of inter-sentential relationships (e.g., semantic similarity \& natural connection), nor model intra-sentential relationships (e.g, semantic \& syntactic relationship among words).
To address these problems, we propose a novel Multiplex Graph Convolutional Network (Multi-GCN) to jointly model different types of relationships among sentences and words.
Based on Multi-GCN, we propose a Multiplex Graph Summarization (Multi-GraS) model for extractive text summarization. 
Finally, we evaluate the proposed models on the CNN/DailyMail benchmark dataset to demonstrate the effectiveness of our method. 
\end{abstract}

\section{Introduction}
Numerous documents from a variety of sources are uploaded to the Internet or database everyday, such as news articles \cite{hermann2015teaching}, scientific papers \cite{qazvinian2008scientific} and electronic health records \cite{jing2019show}.
How to effectively digest the overwhelming information has always been a fundamental question in natural language processing \cite{nenkova2011automatic}.
This question has sparked the research interests in the task of extractive text summarization, which aims to generate a short summary of a document by extracting the most representative sentences from it.

Most of the recent methods \cite{cheng-lapata-2016-neural,narayan-etal-2018-ranking,luo-etal-2019-reading,wang-etal-2020-heterogeneous, mendes2019jointly, zhou2018neural} formulate the task of extractive text summarization as a sequence labeling task, where the labels indicate whether a sentence should be included in the summary. 
To extract sentence features, existing approaches generally use Recurrent Neural Networks (RNN) \cite{yasunaga2017graph, nallapati2017summarunner, zhou2018neural, mendes2019jointly, luo-etal-2019-reading, cheng-lapata-2016-neural}, Convolutional Neural Networks (CNN) \cite{cheng-lapata-2016-neural, luo-etal-2019-reading, narayan-etal-2018-ranking} or Transformers \cite{zhong-etal-2019-searching, liu-lapata-2019-hierarchical}.
Endeavors have been made to develop models to capture various sentence-level relations. 
Early studies, such as LexRank \cite{2004LexRank} and TextRank \cite{mihalcea-tarau-2004-textrank}, built similarity graphs among sentences and leverage PageRank \cite{page1999pagerank} to score them.
Later, graph neural networks e.g., Graph Convolutional Network (GCN) \cite{kipf2016semi} have been adopted on various inter-sentential graphs, such as the approximate discourse graph \cite{yasunaga2017graph}, the discourse graph \cite{xu2020discourse} and the bipartite graph between sentences and words \cite{wang-etal-2020-heterogeneous, jia-etal-2020-neural}.

\begin{figure}
    \centering
    \includegraphics[width=0.45\textwidth]{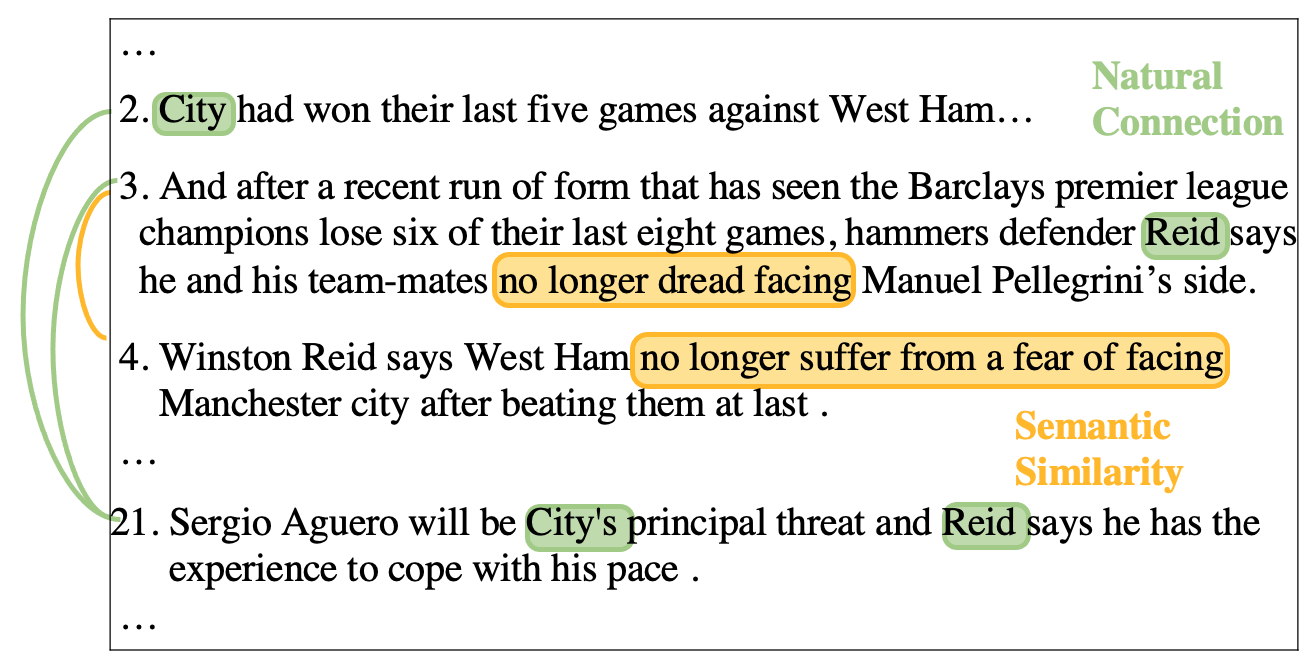}
    \vspace{-0.3cm}
    \caption{An example document:
    There are two different relationships among sentences: the semantic similarity (yellow) and the natural connection (green).
    Sentences 2, 3, 21 are the oracle sentences.}
    \label{fig:example}
    \vspace{-0.4cm}
\end{figure}

Albeit the effectiveness of the existing methods, there are still two under-explored problems.
Firstly, the constructed graphs of the existing studies only involve one type of edges, while sentences are often associated with each other via multiple types of relationships (referred to as the \textit{multiplex graph} in the literature \cite{de2013mathematical, jing2021hdmi}).
Two sentences with some common keywords are considered to be naturally connected (we refer this type of graph as the \textit{natural connection graph}). For example, in Figure \ref{fig:example}, the first and the last sentence exhibit a natural connection (green) via the shared keyword ``City''. 
Although the two sentences are far away from each other, they can be jointly considered as part of the summary since the entire document is about the keyword ``City''.
However, such a relation can barely be captured by traditional encoders such as RNN and CNN.
Two sentences sharing similar meanings are also considered to be connected (we refer this type of graph as the \textit{semantic graph}).
In Figure \ref{fig:example}, the second and the third sentence are semantically similar since they express a similar meaning (yellow).
The semantic similarity graph maps the semantically similar sentences into the same cluster and thus helps the model to select sentences from different clusters, which could improve the coverage of the summary.
Different relationships provide relational information from different aspects, and jointly modeling different types of edges will improve model's performance \cite{wang2019heterogeneous, park2020unsupervised, jing2021network, yan2021dynamic, jing2021mvp}. 
Secondly, the aforementioned methods fall short in taking advantage of the valuable relational information among words.
Both of the syntactic relationship \cite{tai2015improved, he2017deep} and the semantic relationship among words \cite{kenter2015short, wang2020coarse, varelas2005semantic, wang2021cross, radev-etal-2004-mead} have been proven to be useful for the downstream tasks, such as text classification \cite{kenter2015short, jing2018cross}, information retrieval \cite{varelas2005semantic} and text summarization \cite{radev-etal-2004-mead}.

We summarize our contributions as follows: 
\vspace{-0.2cm}
\begin{itemize}
    \setlength\itemsep{-0.1cm}
    \item To exploit multiple types of relationships among sentences and words, we propose a novel Multiplex Graph Convolutional Network (Multi-GCN).
    \item Based on Multi-GCN, we propose a \underline{Multi}plex \underline{Gra}ph based \underline{S}ummarization (Multi-GraS) framework for extractive text summarization.
    \item We evaluate our approach and competing methods on the CNN/DailyMail benchmark dataset and the results demonstrate our models' effectiveness and superiority.
\end{itemize}

\section{Methodology}
We first present Multi-GCN to jointly model different relations, and then present the Multi-GraS approach for extractive text summarization.

\subsection{Multiplex Graph Convolutional Network}\label{sec:multi_gcn}
\begin{figure*}
    \centering
    \begin{subfigure}[b]{.25\textwidth}
        \includegraphics[width=1\textwidth]{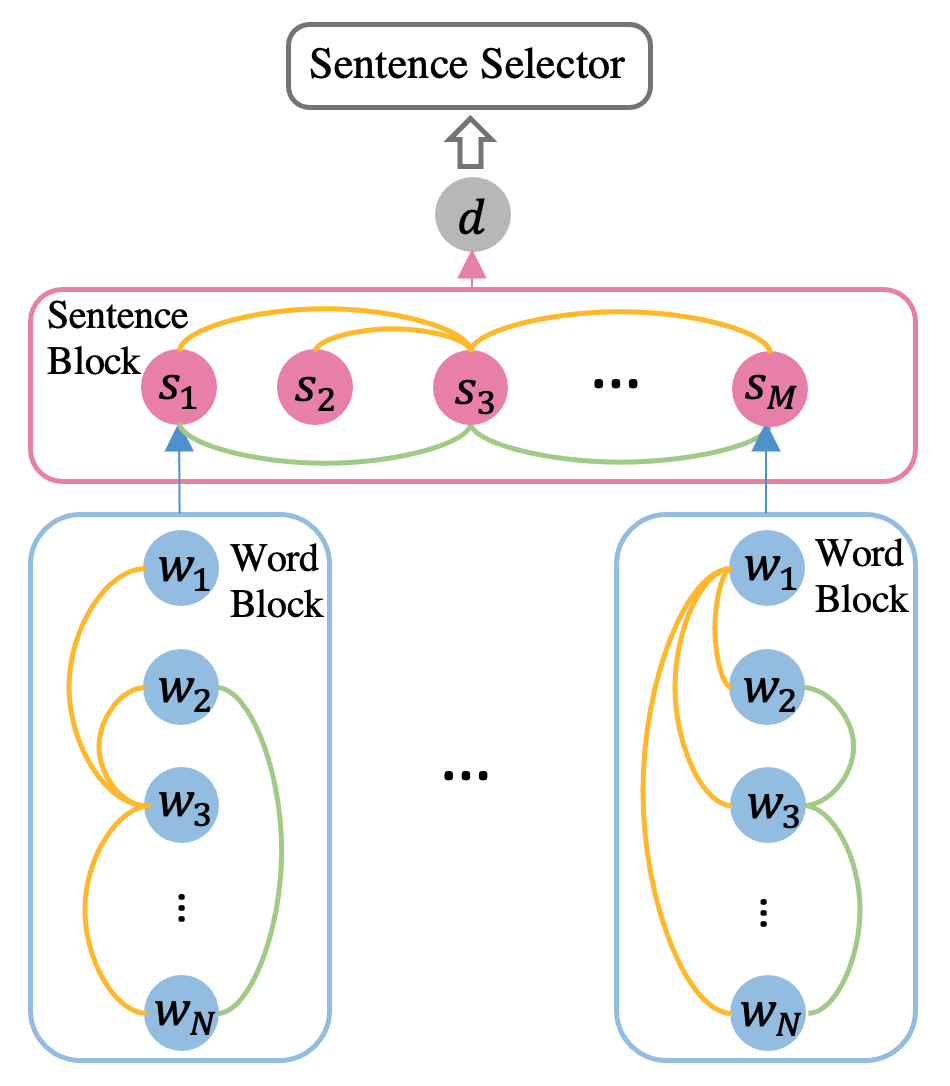}
        \caption{Multi-GraS}\label{fig:overview}
    \end{subfigure}
    \quad
    \begin{subfigure}[b]{.3\textwidth}
        \includegraphics[width=1\textwidth]{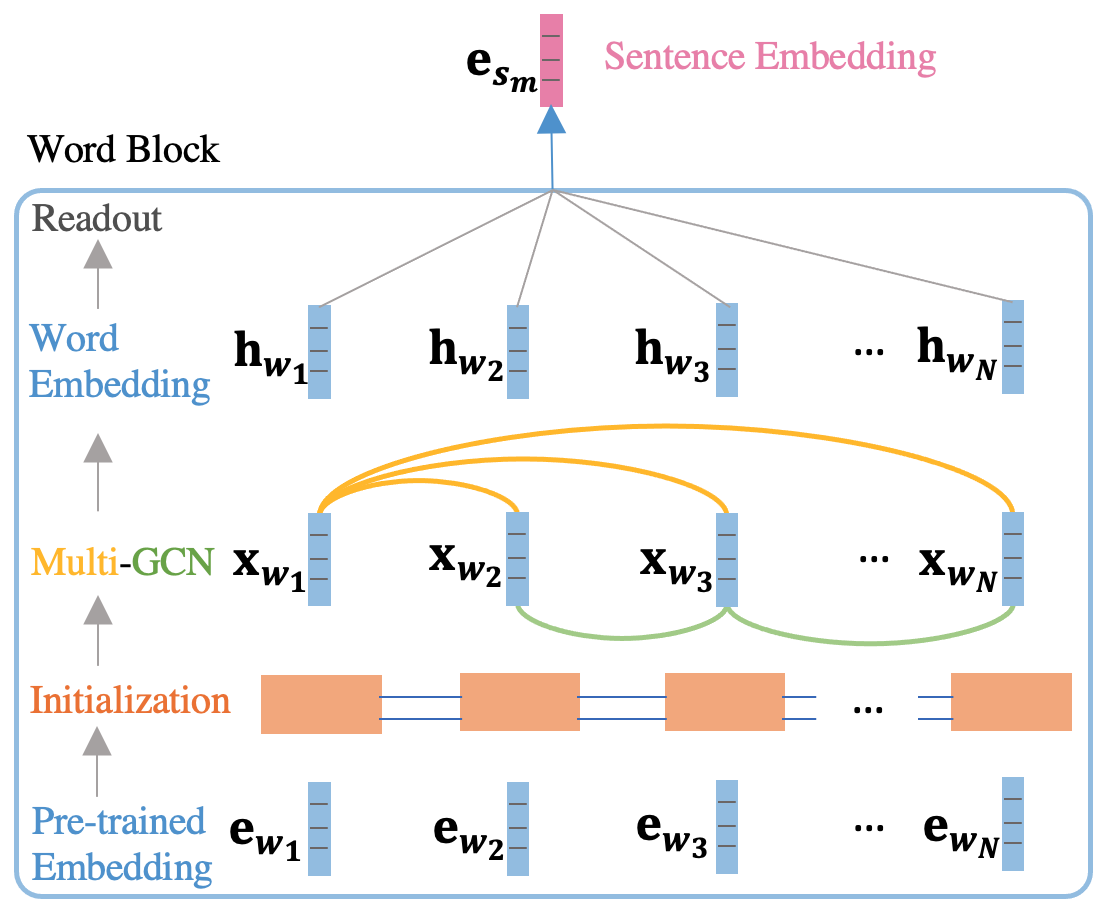}
        \caption{Word Block}\label{fig:word_block}
    \end{subfigure}
    \quad
    \begin{subfigure}[b]{.3\textwidth}
        \includegraphics[width=1\textwidth]{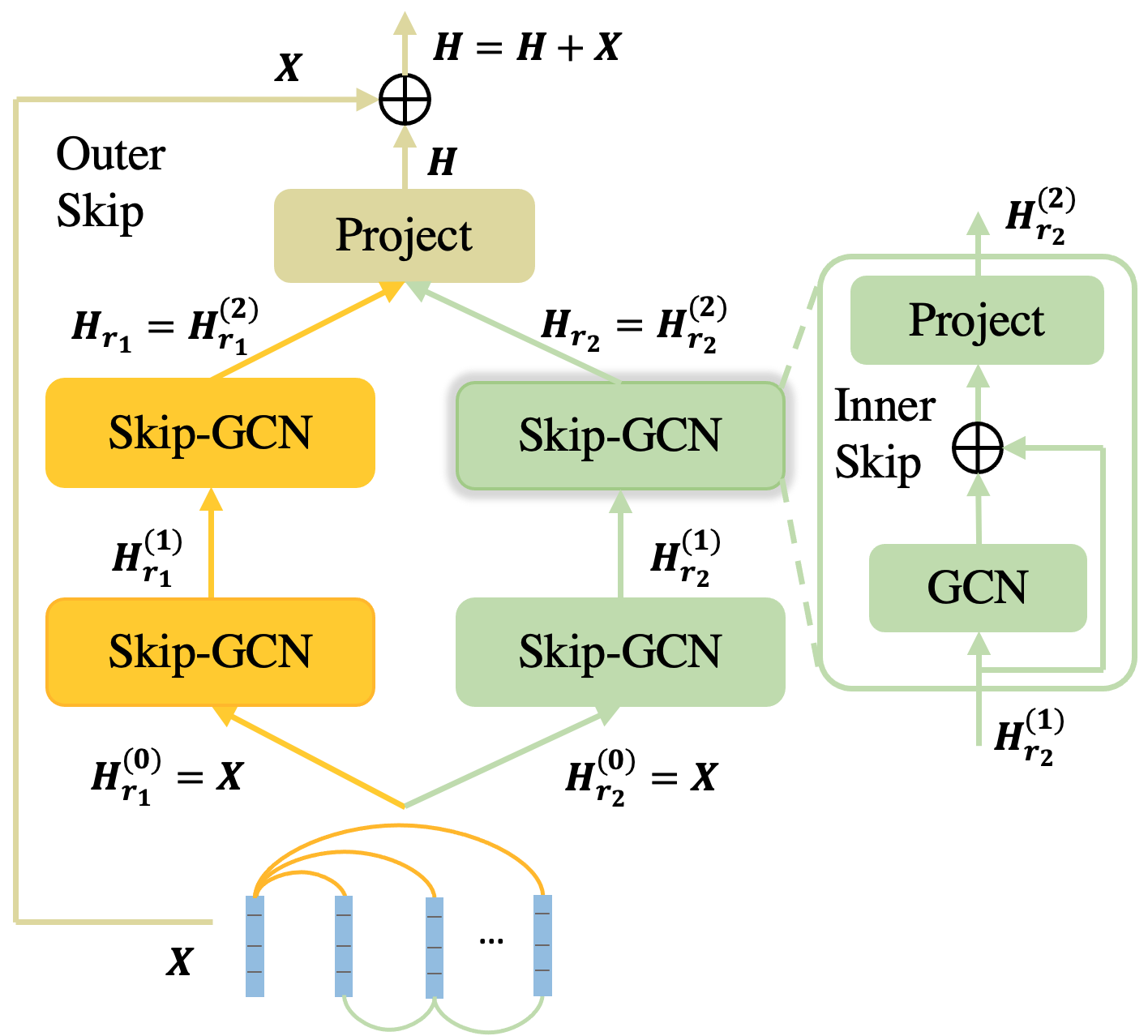}
        \caption{Multi-GCN}\label{fig:multi_gcn}
    \end{subfigure}
    \caption{Overview of the proposed Multi-GraS, the word block and Multi-GCN.}
\end{figure*}

Figure \ref{fig:multi_gcn} illustrates Multi-GCN over a multiplex graph with initial node embedding $\mathbf{X}$ and a set of relations $\mathcal{R}$.
Firstly, Multi-GCN learns node embeddings $\mathbf{H}_r$ of different relations $r\in\mathcal{R}$ separately, and then combines them to produce the final embedding $\mathbf{H}$.
Secondly, Multi-GCN employs two types of skip connections, the inner and the outer skip-connections, to mitigate the over-smoothing \cite{li2018deeper} and the vanishing gradient problems of the original GCN \cite{kipf2016semi}.

More specifically, we propose a Skip-GCN with an inner skip connection to extract the embeddings $\mathbf{H}_r$ for each relation.
The updating functions for the $l$-th layer of the Skip-GCN are defined as: 
\begin{align}
    \hat{\mathbf{H}}_r^{(l)} &= \text{GCN}_r^{(l)}(\mathbf{A}_r, \mathbf{H}_r^{(l-1)}) + \mathbf{H}_r^{(l-1)}\\
    \mathbf{H}_r^{(l)} &= \text{ReLU}(\hat{\mathbf{H}}_r^{(l)}\mathbf{W}^{(l)}_{r} + \mathbf{b}^{(l)}_{r})
\end{align}
where $\mathbf{A}_r$ is the adjacency matrix for the relation $r$;
$\mathbf{W}^{(l)}_{r}$ and $\mathbf{b}^{(l)}_{r}$ denote the weight and bias.
Note that $\mathbf{H}_r^{(0)} = \mathbf{X}$ is the initial embedding, and $\mathbf{H}_r$ is the output after all Skip-GCN layers.

Next, we combine the embedding of different relations $\{\mathbf{H}_{r}\}_{r\in\mathcal{R}}$ by the following equations:
\begin{equation}
    \mathbf{H} = \text{tanh}(\text{cat}(\{\mathbf{H}_{r}\}_{r\in\mathcal{R}})\mathbf{W}+\mathbf{b})
\end{equation}
where cat denotes the concatenation operation and $\mathbf{W}$ and $\mathbf{b}$ denote the weight and bias of the project block in Figure \ref{fig:multi_gcn}.

Finally, we use an outer skip connection to directly connect $\mathbf{X}$ with $\mathbf{H}$:
\begin{equation}
    \mathbf{H} = \mathbf{H} + \mathbf{X}
\end{equation}

\vspace{-0.2cm}
\subsection{The Multi-GraS model}\label{sec:multigras}
The overview of the proposed Multi-GraS is illustrated in Figure \ref{fig:overview}.
Multi-GraS is comprised of three major components: the word block, the sentence block, and the sentence selector.
The word block and the sentence block share a similar ``Initialization -- Multi-GCN -- Readout'' structure to extract the sentence and document embeddings.
The sentence selector picks the most representative sentences as the summary based on the extracted embeddings.

\subsubsection{The Word Block}
The architecture of a word block is illustrated in Figure \ref{fig:word_block}.
Given a sentence $s_m$ with $N$ words $\{w_n\}_{n=1}^N$, the word block takes the pre-trained word embeddings $\{\mathbf{e}_{w_n}\}_{n=1}^N$ as inputs, and produces the sentence embedding $\mathbf{e}_{s_m}$.
Specifically, the \textbf{Initialization} module produces contextualized word embeddings $\{\mathbf{x}_{w_n}\}_{n=1}^N$ via Bi-LSTM.
The \textbf{Multi-GCN} module jointly captures multiple relations for $\{\mathbf{x}_{w_n}\}_{n=1}^N$ and produces $\{\mathbf{h}_{w_n}\}_{n=1}^N$.
The \textbf{Readout} module produces the sentence embedding $\mathbf{e}_{s_m}$ based on max pooling over $\{\mathbf{h}_{w_n}\}_{n=1}^N$.

In this paper, we jointly consider the syntactic and semantic relation among words.
For the syntactic relation, we use a dependency parser to construct a syntactic graph $\mathbf{A}_{syn}$: if a word $w_{n}$ and another word $w_{n'}$ has a dependence link between them, then $\mathbf{A}_{syn}[n, n'] = 1$, otherwise $\mathbf{A}_{syn}[n, n'] = 0$. 
For the semantic relation, we use the absolute value 
of dot product between the embeddings of words to construct the graph: $\mathbf{A}_{sem_w}[n,n'] = |\mathbf{x}_{w_n}^T\cdot\mathbf{x}_{w_{n'}}|$.
Note that we use the absolute value since GCN \cite{kipf2016semi} requires the values in the adjacency matrix to be non-negative.

\subsubsection{The Sentence Block}
Given a document with $M$ sentences $\{s_m\}_{m=1}^M$, the sentence block takes the sentence embeddings $\{\mathbf{e}_{s_m}\}_{m=1}^M$ as inputs, and generates a document embedding $\mathbf{d}$ through a Bi-LSTM, a Multi-GCN and a pooling module.
Essentially, the architecture of the sentence block resembles the word block, thus we only elaborate the construction of graphs for sentences.

In this paper, we consider the natural connection and the semantic relation between sentences.
The semantic similarity between $s_m$ and $s_m'$ is the absolute value of dot product between $\mathbf{x}_{s_m}$ and $\mathbf{x}_{s_{m'}}$, and thus the semantic similarity graph $\mathbf{A}_{sem_s}$ can be constructed by $\mathbf{A}_{sem_s}[m, m']=|\mathbf{x}_{s_m}^T\cdot\mathbf{x}_{s_{m'}}|$.
For the natural connection, if two sentences share a common keyword, then we consider they are naturally connected.
Such a relation helps to cover more sections of a document by connecting far-away sentences (not necessarily semantic similar) via their shared keywords, as shown in Figure \ref{fig:example}.
\begin{equation}
    \mathbf{A}_{nat}[m, m'] = \sum_{w\in\mathcal{W}}\text{tfidf}_{(s_m, w)}\cdot\text{tfidf}_{(s_{m'}, w)},
\end{equation}
where $\text{tfidf}_{(s_m, w)}$ is the tfidf score of the keyword $w$ within $s_m$; $\mathcal{W}$ is the set of keywords.

\subsubsection{Sentence Selector}
The sentence selector first scores the sentences $\{s_m\}_{m=1}^M$ and then selects the top-$K$ sentences as the summary. 
The model design for scoring the sentences follows the human reading comprehension strategy \cite{Pressley1995VerbalPO, luo-etal-2019-reading}, which contains reading and post-reading processes.
The reading process extracts rough meaning of $s_m$:
\begin{equation}
    \mathbf{o}_{m} = \text{tanh}(\mathbf{W}_{Reading}\mathbf{h}_{s_m} + \mathbf{b}_{Reading}).
\end{equation}
The post-reading process further captures the auxiliary contextual information -- document embedding $\mathbf{e}_d$ and the initial sentence embedding $\mathbf{e}_{s_m}$:
\begin{align}
    \mathbf{o}_{m} = \text{tanh}(\mathbf{W}_{Post}[\mathbf{o}_{m}, \mathbf{e}_d, \mathbf{e}_{s_m}] + \mathbf{b}_{Post}).
\end{align}
The final score for $s_m$ is given by:
\begin{equation}
    p_{m} = \sigma(\mathbf{W}_p\mathbf{o}_{m} + \mathbf{b}_p),
\end{equation}
where $\sigma()$ denotes the sigmoid activation.

When ranking the sentences $\{s_m\}_{m=1}^M$, we follow \cite{paulus2018deep, liu-lapata-2019-text, wang-etal-2020-heterogeneous} and use the tri-gram blocking technique to reduce the redundancy. 

\section{Experiments}
\subsection{Experimental Setup}

\subsubsection{Datasets} 

We evaluate our propose model on the benchmark CNN/DailMail \cite{hermann2015teaching} dataset.
This dataset is a combination of the CNN and DailyMail datasets, which contains $287,227$, $13,368$ and $11,490$ articles for training, validating and testing respectively.

For the DailyMail dataset \cite{hermann2015teaching}, the news articles were collected from the DailyMail website.
Each article contains a story and highlights, and the story and highlights are treated as the document and the summary respectively.
The dataset contains $219,506$ articles, which is split into $196,961/12,148/10,397$ for training, validating and testing.

For the CNN dataset \cite{hermann2015teaching}, the news articles were collected from the CNN website.
Each article is comprised of a story and highlights, where the story is treated as the document and highlights are considered as the summary. 
The CNN dataset contains $92,579$ articles in total, $90,266$ are used for training, $1,220$ for validation and $1,093$ for testing.

\subsubsection{Comparison Methods}
For the task of extractive text summarization, we compare the proposed Multi-GraS method with the following methods in three categories: (1) deep learning based methods: NN-SE \cite{cheng-lapata-2016-neural}, LATENT \cite{zhang2018neural}, NeuSUM \cite{zhou2018neural}, JECS \cite{xu2019neural} and EXCONSUMM$_{\text{Extractuve}}$ \cite{mendes2019jointly}; (2) reinforcement-learning based methods: REFRESH \cite{narayan-etal-2018-ranking}, BanditSum \cite{dong2018banditsum}, LSTM+PN+RL \cite{zhong-etal-2019-searching} and HER \cite{luo-etal-2019-reading}; (3) graph based methods: TextRank \cite{mihalcea-tarau-2004-textrank} and HSG \cite{wang-etal-2020-heterogeneous}.

\subsection{Implementation Details}
The vocabulary size is fixed as $50, 000$ and the pre-trained Glove embeddings \cite{pennington2014glove} are used for the input word embeddings. For both of the word block and the sentence block, the Initialization modules employ two-layer Bi-LSTMs.
The Multi-GCN modules use two-layer Skip-GCNs.
We fix all the hidden dimensions as $300$.
We use the Stanford CoreNLP\cite{manning2014stanford} to extract syntactic graphs.
For the natural connection graphs, we filter out the stop words, punctuation, and the words whose document frequency is less than $100$.
During training, we use the Adam optimizer \cite{kingma2014adam}, and the learning rates for CNN, DailyMail, and CNN/DailyMail datasets are set to be $0.0001$, $0.0005$, and $0.0005$, respectively. 
When generating summaries, we select the top-2 and top-3 sentences for the CNN and DailyMail datasets, respectively.

\subsubsection{Oracle Label Extraction}
The summaries of the documents are the highlights of the news written by human experts, hence the sentence-level labels are not provided.
Given a document and its summary, we follow \cite{wang-etal-2020-heterogeneous, liu-lapata-2019-text, mendes2019jointly, narayan-etal-2018-ranking} to identify the set of sentences (or oracles) of the document which has the highest ROUGE scores with respect to its summary.

\subsubsection{Evaluation Metrics}
We evaluate the quality of the summarization by the ROUGE scores \cite{lin2004rouge}, including R-1, R-2 and R-L for calculating the unigram, bigram and the longest common sub-sequence overlapping between the generated summarization and the ground-truth summary. 
In addition to automatic evaluation via ROUGE, we follow \cite{luo-etal-2019-reading, Wu2018LearningTE} and conduct human evaluation to score the quality of the generated summaries.

\begin{table}[t!]
    \scriptsize
    \centering
    \begin{tabular}{l|ccc}
    \hline
    Methods &  R-1 & R-2 & R-L\\
    \hline
    NN-SE & 35.50 & 14.70 & 32.20\\
    LATENT & 41.05 & 18.77 & 37.54 \\
    NeuSUM  & 41.59 & 19.01 & 37.98 \\
    JECS  & 41.70 & 18.50 & 37.90 \\
    EXCONSUMM$_{\text{Extractive}}$ & 41.70 & 18.60 & 37.80 \\\hline
    REFRESH & 40.00 & 18.20 & 36.60 \\
    BanditSum & 41.50 & 18.70 & 37.60 \\
    LSTM+PN+RL & 41.85 & 18.93 & 38.13 \\
    HER   & 42.30 & 18.90 & 37.60 \\\hline
    TextRank & 40.20 & 17.56 & 36.44 \\
    HSG & 42.95 & 19.76 & 39.23 \\\hline
    Multi-GraS & \textbf{43.16} & \textbf{20.14} & \textbf{39.49} \\
    \hline
    \end{tabular}
    \caption{ROUGE scores of different methods.}
    \label{tab:main_results}
\end{table}

\subsection{Overall Performance}
The ROUGE \cite{lin2004rouge} scores of all comparison methods are presented in Table \ref{tab:main_results}.
Within baseline methods, HSG achieves the highest performance, which indicates that considering graph structures could improve performance.
We also observe that Multi-GraS outperforms all of the comparison methods and it achieves $0.21/0.38/0.26$ performance increase on R-1/R-2/R-L scores. 

\begin{table}[t!]
    \scriptsize
    \centering
    \begin{tabular}{l|ccc}
    \hline
    Methods & R-1 & R-2 & R-L \\
    \hline
    Multi-GraS & \textbf{43.16} & \textbf{20.14} & \textbf{39.49} \\
    - trigram blocking & 42.15 & 19.62 & 38.55\\
    - contextual information & 43.12 & 20.04 & 39.18\\
    - outer skip & 43.08 & 20.08 & 39.44 \\
    - inner skip & 42.95 & 20.05 & 39.33\\
    - semantic relation & 43.03 & 20.10 & 39.40\\
    - natural connection relation & 42.74 & 19.80 & 39.14\\
    - weights for natural connection & 42.54 & 19.67 & 38.92 \\
    \hline
    Multi-GraS$_{\text{word}}$ & 42.67 & 19.80 & 39.06\\
    - outer skip & 42.44 & 19.52 & 38.81 \\
    - inner skip & 42.64 & 19.76 & 39.04 \\
    - semantic relation & 42.63 & 19.68 & 39.02\\
    - syntactic relation & 42.42 & 19.57 & 38.82 \\
    \hline
    LSTM & 42.35 & 19.51 & 38.73 \\
    LSTM (w/o tri-gram blocking) & 41.55 & 19.14 & 37.98 \\
    Transformer (w/o tri-gram blocking) & 41.33 & 18.83 & 37.65 \\
    \hline
    \end{tabular}
    \caption{Ablation Study.}
    \label{tab:ablation_study}
\end{table}

\subsection{Ablation Study}\label{sec:abstudy}

Firstly, as shown in Table \ref{tab:ablation_study}, tri-gram blocking and contextual information within the sentence selector help improve model's performance.

Then we study the influence of the Multi-GCN within the sentence block and the word block separately.
To do so, we remove the Multi-GCN within the sentence block (Multi-GraS$_\text{word}$) and further remove the Multi-GCN within the word block (LSTM).
By comparing LSTM, Multi-GraS$_\text{word}$ and Multi-GraS, it can be observed that Multi-GCN in both sentence and word blocks significantly improve the performance.
Next, we study the influence of the components within Multi-GCN.
Table \ref{tab:ablation_study} indicates that the inner and outer skip connections play an important role in Multi-GCN.
Besides, jointly considering different relations is always better than considering one relation alone.

Finally, for the Initialization module in the word and sentence blocks, LSTM performs better than Transformer \cite{vaswani2017attention}. 

\subsection{Human Evaluation}
We randomly select 50 documents along with the summaries obtained by HSG, Multi-GraS, Multi-GraS$_{\text{word}}$, LSTM as well as the oracle summaries.  
Three volunteers (proficiency in English) rank the summaries from $1$ to $5$ in terms of the overall quality, coverage and non-redundancy.
The human evaluation results are presented in Table \ref{tab:human_evaluation}: oracle ranks the highest, Multi-GraS ranks higher than HSG.

\begin{table}[h]
    \scriptsize
    \centering
    \begin{tabular}{l|c|c|c}
    \hline
    Methods & Overall & Coverage & Non-Redundancy \\
    \hline
    LSTM & 3.07 & 2.97 & 2.77 \\
    Multi-GraS$_{\text{word}}$ & 2.97 & 2.93 & 2.77 \\
    HSG & 2.87 & 2.87 & 2.67 \\
    Multi-GraS & 2.20 & 2.23 & 2.13 \\
    Oracle & 1.70 & 1.57 & 1.57 \\
    \hline
    \end{tabular}
    \caption{Human evaluation: the lower the better.}
    \label{tab:human_evaluation}
\end{table}

\section{Sensitivity Experiments}
\begin{figure}
    \centering
    \begin{subfigure}[b]{.23\textwidth}
        \includegraphics[width=1\textwidth]{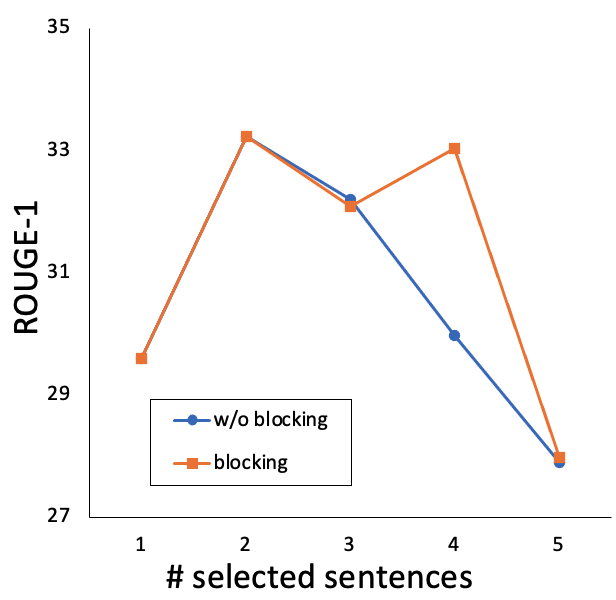}
        \caption{CNN}
    \end{subfigure}
    \begin{subfigure}[b]{.23\textwidth}
        \includegraphics[width=1\textwidth]{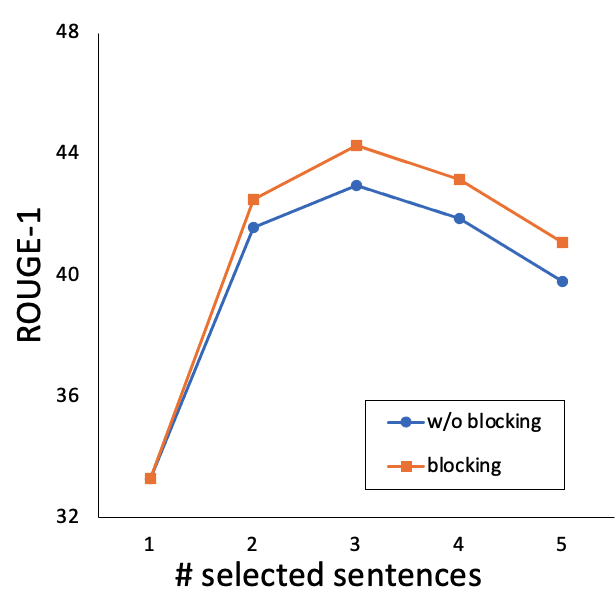}
        \caption{DailyMail}
    \end{subfigure}
    \caption{Rouge-1 score vs. the number of selected sentences.}
    \label{fig:sensitivity}
\end{figure}

To check the performance on the number of selected sentences, we conduct a sensitivity experiment for both CNN and DailyMail datasets. 
The results in Figure \ref{fig:sensitivity} show that the Multi-GraS performs the best when the number of the selected sentences is $2$ for the CNN dataset and $3$ for the DailyMail dataset.

\section{Conclusion}
In this paper, we propose a novel Multi-GCN to jointly model multiple relationships among words and sentences.
Based on Multi-GCN, we propose a novel Multi-GraS model for extractive text summarization.
Experimental results on the benchmark CNN/DailyMail dataset demonstrate the effectiveness of the proposed methods.

\section*{Acknowledgements}
Jing and Tong are partially supported by NSF (1947135 and 1939725).

\bibliography{anthology}
\bibliographystyle{acl_natbib}
\end{document}